\documentclass[runningheads]{llncs}
\usepackage{floatrow}
\usepackage{graphicx}
\usepackage{amsmath,amssymb} % define this before the line numbering.
\usepackage{algorithm}% http://ctan.org/pkg/algorithms
\usepackage{algpseudocode}% http://ctan.org/pkg/algorithmicx
\usepackage{amsmath,amssymb} % define this before the line numbering.
\usepackage{color}
\usepackage{csquotes}
\usepackage{comment}
\usepackage{subfigure}
\DeclareMathOperator*{\argmax}{argmax} % thin space, limits underneath in displays
\DeclareMathOperator*{\argmin}{argmin}
\usepackage{floatrow}
\newfloatcommand{capbtabbox}{table}[\captop][\FBwidth]

\begin{document}
% \linenumbers
\pagestyle{headings}
\mainmatter
\def\ECCVSubNumber{3243}  % Insert your submission number here

%\title{PONAS: Progressive One-shot Neural Architecture Search for Very Efficient Deployment} % Replace with your title
\title{PONAS: Progressive One-shot Neural Architecture Search for Very Efficient Deployment}

\begin{comment}
\titlerunning{ECCV-20 submission ID \ECCVSubNumber} 
\authorrunning{ECCV-20 submission ID \ECCVSubNumber} 
\author{Anonymous ECCV submission}
\institute{Paper ID \ECCVSubNumber}
\end{comment}
%******************

% \begin{comment}
\titlerunning{PONAS}
\author{Sian-Yao Huang \and
Wei-Ta Chu}
\authorrunning{Sian-Yao Huang ,
Wei-Ta Chu}
\institute{National Cheng Kung University, Tainan, Taiwan\\
\email{\{P76084245,10808027\}@gs.ncku.edu.tw}}
% \end{comment}
%******************
\maketitle

\begin{abstract}
We achieve very efficient deep learning model deployment that designs neural network architectures to fit different hardware constraints. Given a constraint, most neural architecture search (NAS) methods either sample a set of sub-networks according to a pre-trained accuracy predictor, or adopt the evolutionary algorithm to evolve specialized networks from the supernet. Both approaches are time consuming. Here our key idea for very efficient deployment is, when searching the architecture space, constructing a table that stores the validation accuracy of all candidate blocks at all layers. For a stricter hardware constraint, the architecture of a specialized network can be very efficiently determined based on this table by picking the best candidate blocks that yield the least accuracy loss. To accomplish this idea, we propose Progressive One-shot Neural Architecture Search (PONAS) that combines advantages of progressive NAS and one-shot methods. In PONAS, we propose a two-stage training scheme, including the meta training stage and the fine-tuning stage, to make the search process efficient and stable. During search, we evaluate candidate blocks in different layers and construct the accuracy table that is to be used in deployment. Comprehensive experiments verify that PONAS is extremely flexible, and is able to find architecture of a specialized network in around 10 seconds. In ImageNet classification, 75.2\% top-1 accuracy can be obtained, which is comparable with the state of the arts. 

\end{abstract}

\section{Introduction}
Deep neural networks have brought surprising advances in various research fields. Promising performance can be achieved if the networks are well designed and trained based on large amounts of data. However, designing good neural networks specific to a collection of constraints requires much domain knowledge, rich experience on model training and tuning, and a lot of time on trials and errors. Neural architecture search (NAS) is thus important and urgently-demanded to automate model design. Generally NAS methods can be categorized according to three dimensions \cite{elsken19}: search space, search strategy, and performance estimation strategy. There have been a dozen of studies proposed to search for neural architectures, especially for the task of image recognition. However, they are mostly time-consuming because of the intractable search space or expensive search strategy. Moreover, most of them are not scalable to various hardware constraints so that specialized networks should be determined and trained for each case. 

Architecture searching is computationally expensive. For example, typical NAS methods based on reinforcement learning (RL) requires tens of GPU days \cite{baker17}\cite{zoph17}. The recent RL-based NAS method, e.g., MnasNet \cite{Tan_2019_CVPR}, was estimated to require about $40K$ GPU hours for one specific network \cite{cai2018proxylessnas}. 

To reduce search efforts, dfferentiable NAS (DNAS) methods and one-shot NAS methods emerged recently. They both can be viewed as \emph{weight sharing} approaches. Conceptually, DNAS models the search space as an architecture distribution described by architecture parameters. The architecture distribution is embodied by learning a supernet that is described by supernet weights. Architecture search and model training are tightly coupled. After the supernet is constructed, the optimal architectures are sampled from the trained distribution. However, the network sampled from the architecture distribution is only suitable to a specific hardware constraint. This makes DNAS less scalable. The FBNet \cite{Wu_2019_CVPR} is a differentiable NAS framework that largely speeds up searching for a specific network in $216$ GPU hours. But if we need $N$ different networks specific to $N$ different constraints, $216 \times N$ GPU hours are still needed.

In contrast to DNAS, one-shot NAS methods \cite{1904.00420}\cite{cai2020once}\cite{pmlr-v80-bender18a} decouple model training from architecture search. A supernet, or once-for-all (OFA) network \cite{cai2020once}, is trained to flexibly support different sub-networks with different depths, widths, kernel sizes, and resolutions. Given hardware and/or latency constraints, a subset of sub-networks are sampled according to a pre-trained accuracy predictor \cite{cai2020once}, or are evolved through an evolutionary algorithm \cite{1904.00420}. %The OFA network or supernet only need to be trained once, and various sub-networks can be sampled or evolved according to different constraints.
However, training an accuracy predictor is still expensive (40 GPU hours mentioned in \cite{cai2020once}). The time needed to execute the evolutionary algorithm \cite{1904.00420} is considerable, too. 

We conclude two problems in previous weight sharing NAS approaches:
\begin{itemize}
\item High cost of supernet training: Training a supernet needs a lot of computation resource and search time. 
\item High cost of sub-network specialization: No matter DNAS or one-shot NAS, considerable time is needed to do network specialization. 
\end{itemize}

In this work, we propose \emph{Progressive One-shot Neural Architecture Search} (PONAS) that combines the advantages of progressive NAS and the one-shot method. Progressive NASs like \cite{Liu_2018_ECCV} and \cite{Dong_2018_ECCV} construct convolutional neural networks (CNNs) by stacking some predefined numbers of \enquote{cells}. The best structure of the cell is searched by progressively expanding blocks (operations). The determined best cell then acts as a \enquote{layer}. The structure of each layer, therefore, keeps the same. In the proposed PONAS, instead of searching the best cell, we search the best block for each layer progressively. In this way, structures of different layers may be different, and richer expressivity may be obtained. To tackle the first problem mentioned above, we propose a two-stage training scheme that separates the searching process into the meta training stage and the fine-tuning stage to make the search process more efficient and stable. 

In the progressive search process, we construct an \emph{accuracy table} that stores validation accuracy of each candidate block in each layer. Given a hardware constraint, a specialized network can be evolved by the evolutionary algorithm. This evolution is extremely efficient because only simple table lookup is needed to estimate performance of any specialized network. In our experiment, architecture of a specialized network can be determined in around 10 seconds. This approach largely resolves the second problem mentioned above.  

Notice that, in contrast to FBNet \cite{Wu_2019_CVPR} where sub-networks are sampled from a supernet, we directly derive a network specific to the hardware constraint according to the information obtained during the search process of PONAS. This strategy enables us to get multiple specific networks very efficiently based on the accuracy table that only needs to be constructed once. 

\section{Related Works}
%Generally, NAS methods consist of three components: 1) architecture search space, 2) search strategy, and 3) performance estimation strategy \cite{elsken19}. 
Because the formulation of NAS is similar to reinfocement learning (RL), early NAS methods were firstly proposed based on it. However, such approaches are very computationally expensive. A variety of methods were thus proposed based on progressive learning or weight sharing to reduce computational cost. In the following, we only focus on related works on these two approaches. 

\subsection{Progressive Neural Architecture Search}
Progressive NAS (PNAS) methods \cite{Dong_2018_ECCV}\cite{Liu_2018_ECCV} search the architecture space in a progressive way. A sequential model-based optimization (SMBO) strategy \cite{hutter11} is adopted to search for architectures in the order of increasing complexity, while a surrogate model is learnt simultaneously to guide the search. Starting from the first cell, all possible block structures are trained and evaluated. Each of them is then expanded by adding all possible blocks, which largely enlarges the search space. To reduce search time, a performance predictor is trained to evaluate all these extensions, and then only the top blocks are retained. According to \cite{Liu_2018_ECCV}, this approach is about 8 times faster than the RL-based method \cite{zoph18} in terms of total compute.  

Inspired by PNAS, we also adopt the idea of progressive learning. But different from the SMBO strategy, we construct an accuracy table rather than building a surrogate model as a performance predictor. 

\subsection{Weight Sharing Neural Architecture Search}
Many recent NAS approaches are conceptually based on weight sharing \cite{cai2018proxylessnas} \cite{liu2018darts} \cite{Wu_2019_CVPR} \cite{chu2019fairnas} \cite{Yan_2019_ICCV_Workshops}. The main idea is constructing a supernet to represent the entire search space. Such methods can be generally divided into two categories: differentiable NAS and one-shot NAS.

Differentiable NAS \cite{Wu_2019_CVPR}\cite{cai2018proxylessnas}\cite{liu2018darts} views the search space as an architecture distribution described by architecture parameters. Architecture parameters are optimized when training a supernet. After supernet training, the optimal architectures are sampled from the trained distribution. Instead of searching over a discrete set of candidate architectures, Liu et al. \cite{liu2018darts} relaxed the search space from discrete to continuous, and thus the gradient descent algorithm can be adopted to find the optimal architecture. Cai et al. \cite{cai2018proxylessnas} binarized the architecture parameters and forced only one path to be active when training the supernet, which reduces the required GPU memory. %At each update step of architecture parameters, instead of activating all paths to train architecture parameters, they only sampled two paths and updated the corresponding architecture parameters by the gradient descent algorithm. 
Wu et al.\cite{Wu_2019_CVPR} used the Gumbel softmax technique \cite{1611.01144} to find the optimal distribution of architecture parameters.

One-shot NAS \cite{pmlr-v80-bender18a}\cite{1904.00420}\cite{cai2020once}\cite{chu2019fairnas} considers the trained supernet as an evaluator to predict performance of all sub-networks. After training a supernet, one-shot NAS can adopt random sampling, evolutionary algorithms, or reinforcement learning to derive multiple specific networks conforming to different constraints without retraining the supernet. Using the trained supernet as the performance evaluator, Guo et al. \cite{1904.00420} used an evolutionary algorithm to find the optimal architecture for the given constraint. Cai et al. \cite{cai2020once} sampled a set of sub-networks to train an accuracy predictor. This guides architecture search to get a specialized network. Chu et al. \cite{chu2019fairnas} pointed out the problem of biased evaluation, which is prone to misjudgments of candidate architectures. They thus proposed to fairly train candidate blocks to get a supernet as a reliable performance evaluator. 

\section{Progressive One-Shot Neural Architecture Search}
% 透過什麼方式來實做Progressive One-Shot Neural Architecture Search
Progressive NAS \cite{Dong_2018_ECCV} \cite{Liu_2018_ECCV} adopted the SMBO strategy to search for the best structure of the cell by progressively expanding blocks (operations). Inspired by progressive search, the proposed PONAS also searches the architecture progressively, but PONAS directly searches for the best network layer by layer. More importantly, the accuracy table constructed in PONAS process facilitates efficient network specialization without a surrogate model for accuracy prediction. 

\subsection{Overview}
\label{sec:overview}
\subsubsection{Previous Network Specialization.}
Denote the weights of a supernet $A$ as $W_A$. Each sampled architecture $a$ inherits weights from $W_A$. Given a constraint $C$, previous one-shot NAS methods \cite{1904.00420}\cite{chu2019fairnas} achieved network specialization by finding the best sub-network $a^*$ that yields the highest accuracy and fits the constraint $C$. That is,  
\begin{equation}
\begin{alignedat}{2}
a^* = \argmax_{a \in A}ACC_{val}(W_A, a), \\
\text{s.t.} \quad Cost(a^*) \leq C, 
\end{alignedat}
\label{eq:previous}
\end{equation}
where $ACC_{val}(W_A, a)$ is the validation accuracy yielded by the architecture $a$. The network deployment process mentioned in Eqn.~(\ref{eq:previous}) is time consuming because the validation accuracy of sub-networks $a$'s should be calculated case by case. 

\subsubsection{The Proposed Process.}
Instead of calculating validation accuracy in each deployment, we propose to build an accuracy table when constructing the supernet. With this table, very efficient network deployment can be achieved because only simple table lookup is needed. 

Denote the weights of the $l$-th layer as $W^{(l)}$, and the $i$-th candidate block in the $l$-th layer as $B_i^{(l)}$. A candidate block is a set of settings of operations, which may include different kernel sizes, expansion settings, and so on. When constructing the supernet, we would like to find the architecture as a stack of blocks that yields the highest accuracy. For the $l$-th layer, the best block is determined by  
\begin{equation}
    % \argmax_{i \in I}\sum^L_lACC_{val}(W_{\hat{A}_l}, a_{B^{(i)}_l})
    i^* = \argmax_{i = 1, 2, ..., I} ACC_{val}(W_{\hat{A}^{(l)}}, a_{B_i^{(l)}}), 
\end{equation}
where $\hat{A}^{(l)}$ is a supernet containing all candidate blocks in the $l$-th layer, while containing the default block in all other layers (details will be given in Sec.~\ref{sec:twostage}). The term $a_{B_i^{(l)}}$ is a sub-network taking the block $B_i^{(l)}$ in the $l$-th layer, which is sampled from $\hat{A}^{(l)}$. The term $I$ is the total number of candidate blocks in a layer. By finding the best blocks $B_{i^*}^{(1)}$, $B_{i^*}^{(2)}$, ..., $B_{i^*}^{(L)}$ in layers 1 to $L$ progressively, we can stack them to construct the supernet, denoted as $[B_{i^*}^{(1)} \rightarrow B_{i^*}^{(2)} \rightarrow \cdots \rightarrow B_{i^*}^{(L)}]$. 

When finding the best architecture of the supernet, we simultaneously construct the accuracy table $T$ that stores the validation accuracy of all candidate blocks at all layers:
\begin{equation}
    T[l, i] = ACC_{val}(W_{\hat{A}^{(l)}}, a_{B_i^{(l)}}), \quad l = 1, ..., L; \quad i = 1, ..., I. 
\end{equation}

Given a constraint $C$, we can get the specific network $a^*$ that maximizes the accuracy and fits the constraint $C$ by checking the accuracy values stored in $T$. 

\begin{equation}
\begin{alignedat}{2}
\mathcal{I}^* = \argmax_{i \in \mathcal{I}} T(l, i), \\
\text{s.t.} \quad Cost(a_{\mathcal{I}^*}) \leq C, 
\end{alignedat}
\label{eq:findbest}
\end{equation}
where $\mathcal{I}$ is the set of indices of candidate blocks in all layers. The meaning of Eqn.~(\ref{eq:findbest}) is that we want to find the best sequence of blocks $\mathcal{I}^* = (i^{(1)}, i^{(2)},...,i^{(L)})$ among all combinations such that the architecture $a_{\mathcal{I}^*}$ yields the highest accuracy and fits the constraint $C$ simultaneously. The term $i^{(j)}$ denotes the index of the candidate block $B_{i^{(j)}}^{(j)}$ at the $j$-th layer. 

The main difference between Eqn.~(\ref{eq:findbest}) and Eqn.~(\ref{eq:previous}) is that we do not need to calculate or estimate validation accuracy on the fly, but just need to look at the accuracy table. We adopt the genetic algorithm to find the best sequence among all combinations. Because only table lookup is needed, finding the best sub-network architecture usually can be done in 10 seconds. 

% =================================================================================
 
\begin{table}[]
\centering
\scriptsize
\caption{Macro-architecture of the search space. MBConv E1 $3\times 3$ denotes MBConv with kernel size 3 and expansion 1. "Repeat" denotes the number of layers repeat with the corresponding settings.}
\begin{tabular}{|c|c|c|c|c|}
\hline
Input shape       & Block                 & Output channel & Repeat & Stride \\ \hline
$224^2 \times 3$  & Conv 3 $\times$ 3     & 32             & 1      & 2      \\
$112^2 \times 32$ & MBConv E1 $3\times 3$ & 16             & 1      & 1      \\ \hline
$112^2 \times 16$ & Candidate Block       & 32             & 1      & 2      \\
$56^2 \times 32$  & Candidate Block       & 32             & 1      & 1      \\
$56^2 \times 32$  & Candidate Block       & 40             & 1      & 2      \\
$28^2 \times 40$  & Candidate Block       & 40             & 3      & 1      \\
$28^2 \times 40$  & Candidate Block       & 80             & 1      & 2      \\
$14^2 \times 80$  & Candidate Block       & 96             & 4      & 1      \\
$14^2 \times 96$  & Candidate Block       & 96             & 3      & 1      \\
$14^2 \times 96$  & Candidate Block       & 192            & 1      & 2      \\
$7^2 \times 192$  & Candidate Block       & 320            & 4      & 1      \\
$7^2 \times 320$  & Candidate Block       & 1280           & 1      & 1      \\ \hline
$7^2 \times 1280$ & Avg pool $7 \times 7$ & -              & 1      & 1      \\
$1280$            & Fully Connected       & 1000           & 1      & -      \\ \hline
\end{tabular}
\label{table:macro-architecture}
\end{table}

% 基於moblinetv2的inverted residual block
In this work, the setting of the architecture search space is inspired by \cite{cai2018proxylessnas}. The backbone of candidate block is mobile inverted bottleneck convolution (MBConv) \cite{Sandler_2018_CVPR} with kernel sizes \{3,5,7\} and expansion \{3,6\} in depthwise convolution. The squeeze-and-excite \cite{hu2018senet} module is considered to be employed or not to expand the search space. Therefore, we have $I = 3 \times 2 \times 2 = 12$ types of candidate blocks for each layer. The number of layers to be constructed is set to $L = 19$. Accordingly, our search space has a size of $12^{19}$ in total. Particularly, the macro-architecture of the search space is illustrated in Table~\ref{table:macro-architecture}.

\subsection{Two-stage Training}
\label{sec:twostage}
\begin{figure}[t]
    \centering
    \subfigure[]{
        \begin{minipage}[t]{0.28\linewidth}
        \centering
         \raisebox{-0.5\height}{\includegraphics[width=1.2in]{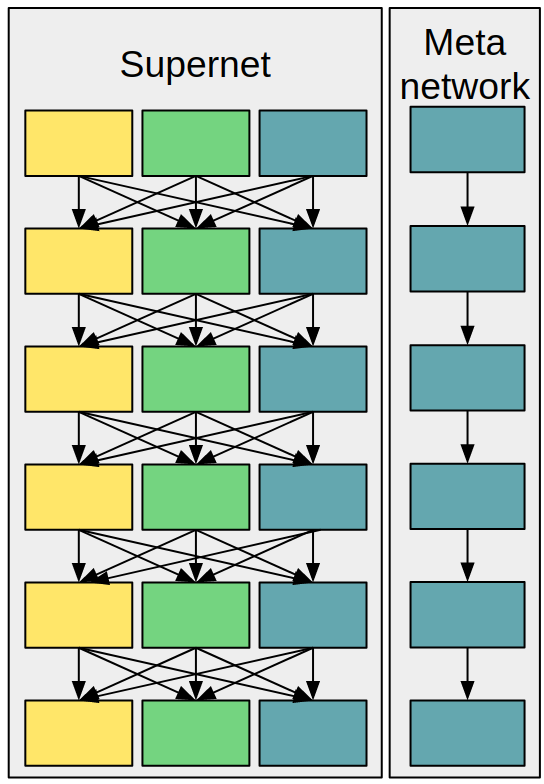}}
        \end{minipage}%
    }%
    \subfigure[]{
        \begin{minipage}[t]{0.72\linewidth}
        \centering
         \raisebox{-0.5\height}{\includegraphics[width=3in]{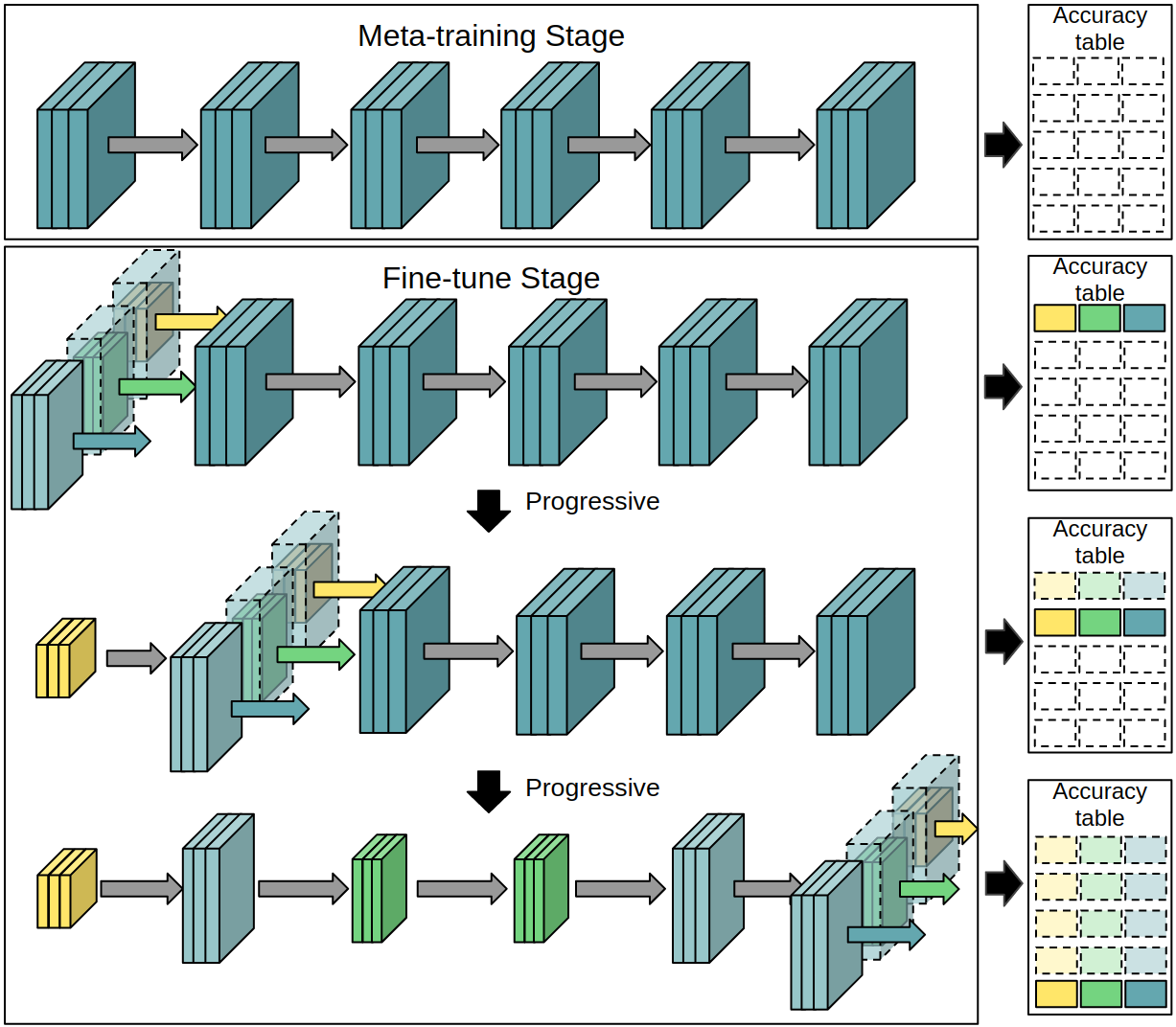}}
        \end{minipage}%
    }%
    \caption{\textbf{(a)} The main difference between a supernet and the proposed meta network is that the former consists of multiple paths of blocks, and the later is single-path. \textbf{(b)} Our two-stage training scheme. In the meta training stage, we only train meta network which is constructed by the largest candidate block. In the fine-tuning stage, we progressively fine-tune the meta network to construct the accuracy table. }
    \label{fig:supernet}
    \centering
\end{figure}

Weight sharing approaches like \cite{Wu_2019_CVPR}\cite{1904.00420}\cite{chu2019fairnas}\cite{liu2018darts} and our approach mentioned in Sec.~\ref{sec:overview} need to construct a generic network (supernet), and then sub-networks are sampled or derived from it to conform to constraints. However, training a supernet needs a lot of computation resource and search time. In \cite{Wu_2019_CVPR}, for example, constructing the supernet needs to consider all combinations of candidate blocks in the search space, as illustrated in the left of Fig.~\ref{fig:supernet}(a). This makes the structure of the supernet quite complex and thus much resource is required. Some recent works \cite{stamoulis2019singlepath}\cite{1912.04749} use the super kernel to encode all candidate blocks into one block, and search the best distribution of architecture parameters to get the best sub-network. But these approaches give rise to the coupling problem mentioned in \cite{1904.00420}. 

To reduce the cost of training a supernet, we propose a two-stage training scheme. This training scheme is compatible to be integrated with other weight sharing NASs. The first stage of this scheme is called \emph{meta training stage}, and the second stage is called \emph{fine-tuning stage}.

In the meta training stage, we construct a meta network which is the \emph{largest network} in the search space. In each layer of the so-called largest network, only the candidate block with the largest convolution kernel, expansion, and enabled squeeze-and-excite module is used, i.e., kernel size = 7, expansion = 6, with squeeze-and-excite module enabled. Let $B_{G}$ denote the largest block in the following. Fig.~\ref{fig:supernet}(a) illustrates the difference between a supernet and the proposed meta network. The meta network is a single path consisting of the concatenation of $B_{G}$'s, and a supernet is a network consisting of multiple paths of various blocks. Training the meta network is thus easier. 

Training the meta network acts as finding good initialization parameters based on the largest block. To further elaborate the network, we propose to progressively fine-tune the meta network by finding the \emph{best block} for each layer. Instead of constructing the entire supernet as the left of Fig.~\ref{fig:supernet}(a), we only replace $B_{G}$ of one layer by 12 candidate blocks each time to construct the supernet $\hat{A}^{(l)}$. For example, to fine-tune the first layer, the default block $B_{G}$ is replaced by candidate blocks $(B_{1}, B_{2}, ..., B_{12})$, and the 2nd layer to the 19th layer keep using the default block $B_{G}$ (with the parameters discovered in the meta learning stage) to construct $\hat{A}^{(1)}$. The supernet $\hat{A}^{(1)}$ is fine-tuned based on the training data with strict fairness \cite{chu2019fairnas}, respectively. Let $[B_{i} \rightarrow B_{G} \rightarrow \cdots \rightarrow B_{G}]$ denote the specific network from $\hat{A}^{(1)}$ where the block in the first layer is $B_i$, followed by $B_{G}$'s. The validation accuracy of each specific network is evaluated and stored in the \emph{accuracy table} $T[1, 1:12]$, which will play an important role in network specialization described later. 

Assume that when $B_{G}$ replaced by the candidate block $B_*^{(1)} \in \{B_{1}, B_{2}, ..., B_{12} \}$, the specific network yields the highest validation accuracy. This network is denoted as $[B_*^{(1)} \rightarrow B_{G}^{(2)} \rightarrow \cdots \rightarrow B_{G}^{(19)}]$. When we try to fine-tune the second layer, parameters of the supernet $\hat{A}^{(2)}$, $[B_*^{(1)} \rightarrow B_{1}^{(2)} \rightarrow \cdots \rightarrow B_{G}^{(19)}]$, $[B_*^{(1)} \rightarrow B_{2}^{(2)} \rightarrow \cdots \rightarrow B_{G}^{(19)}]$, ..., $[B_*^{(1)} \rightarrow B_{12}^{(2)} \rightarrow \cdots \rightarrow B_{G}^{(19)}]$, are updated based on the training data. In the same way, the validation accuracy of each specific network from $\hat{A}^{(2)}$ is evaluated and stored in the accuracy table $T[2, 1:12]$. 

We progressively find the best candidate block for each layer, and store all validation accuracy values in the accuracy table $T$. The progressive fine-tuning process is illustrated in Fig.~\ref{fig:supernet}(b). After fine-tuning all layers, we finally get the network $[B_*^{(1)} \rightarrow B_*^{(2)} \rightarrow \cdots \rightarrow B_{*}^{(19)}]$ and the accuracy table $T$. 

The parameters of \enquote{smaller} candidate blocks $(B_{1}, B_{2}, ..., B_{12})$ are initialized by the parameter of the largest block $B_{G}$. Fig.~\ref{fig:supernet}(b) illustrates how we crop parameters of the largest block $B_{G}$ to get the parameters of all smaller blocks. With such initialization, we just need to fine-tune for a few epochs rather than lots of epochs from random initialization. This idea is similar to progressive shrinking mentioned in \cite{cai2020once}.

\subsection{Network Specialization in the Accuracy Loss Domain}
\label{sec:accuracyloss}
Given a constraint, previous one-shot methods derived a specific network based on the evolutionary algorithm \cite{1904.00420}\cite{cai2020once}. These methods flexibly support different constraints and only need to train the supernet once. However, the evolutionary method proposed in Guo et al.\cite{1904.00420} required much time in measuring validation accuracy of the population, and the method proposed in Cai et al.\cite{cai2020once} needed to train an accuracy predictor to measure validation accuracy of 16K sub-networks. In our work, we employ the accuracy table $T$ as the performance evaluator in the network specialization process. 

A problem from Eqn.~(\ref{eq:findbest}) arises when comparing with the validation accuracy of candidate blocks in different layers. Candidate blocks in different layers are not directly comparable because different levels of information is learnt. Therefore, we argue that they should be compared in a domain commonly for different layers. Chu et al. \cite{chu2019fairnas} pointed out that \enquote{different choice blocks of the same layer learn similar feature maps on the corresponding channel}. Inspired by this observation, we propose to represent performance of each candidate block as \emph{the accuracy loss from the best block at the corresponding layer}. For example, let $t_1^{(l)}, t_2^{(l)}, ..., t_{12}^{(l)} $ be the validation accuracy (evaluated in the training process and stored in the accuracy table $T$) of the candidate blocks $B_1^{(l)}, B_2^{(l)}, ..., B_{12}^{(l)}$ at the $l$-th layer, and let $t_*^{(l)}$ denote the best accuracy yielded by $B_*^{(l)}$, the accuracy loss is calculated as $\Delta \boldsymbol{t}^{(l)} = (t_1^{(l)} - t_*^{(l)}, t_2^{(l)} - t_*^{(l)} , ..., t_{12}^{(l)} - t_*^{(l)} ) = ( \Delta t_1^{(l)}, \Delta t_2^{(l)}, ..., \Delta t_{12}^{(l)} )$. In this way, we calculate $\Delta \boldsymbol{t}^{(1)}, ..., \Delta \boldsymbol{t}^{(19)}$, and transform the performance of candidate blocks in different layers into the \emph{accuracy loss domain}.

We then can construct the accuracy loss table $T_d[l, :] = [\Delta \boldsymbol{t}^{(1)}; \Delta \boldsymbol{t}^{(2)}; ...; \Delta \boldsymbol{t}^{(L)}]$. Based on $T_d$, we make an assumption that \emph{the accuracy loss of candidate blocks can quantify importance of candidate blocks in different layers}. With the accuracy loss table $T_d$, we can reformulate Eqn.~(\ref{eq:findbest}) as
\begin{equation}
\begin{alignedat}{2}
\mathcal{I}^* = \argmin_{i \in \mathcal{I}} T_d(l, i), \\
\text{s.t.} \quad Cost(a_{\mathcal{I}^*}) \leq C. 
\label{eq:ga}
\end{alignedat}
\end{equation}

\begin{figure}[t]
	\centering
	\includegraphics[width=3.3in]{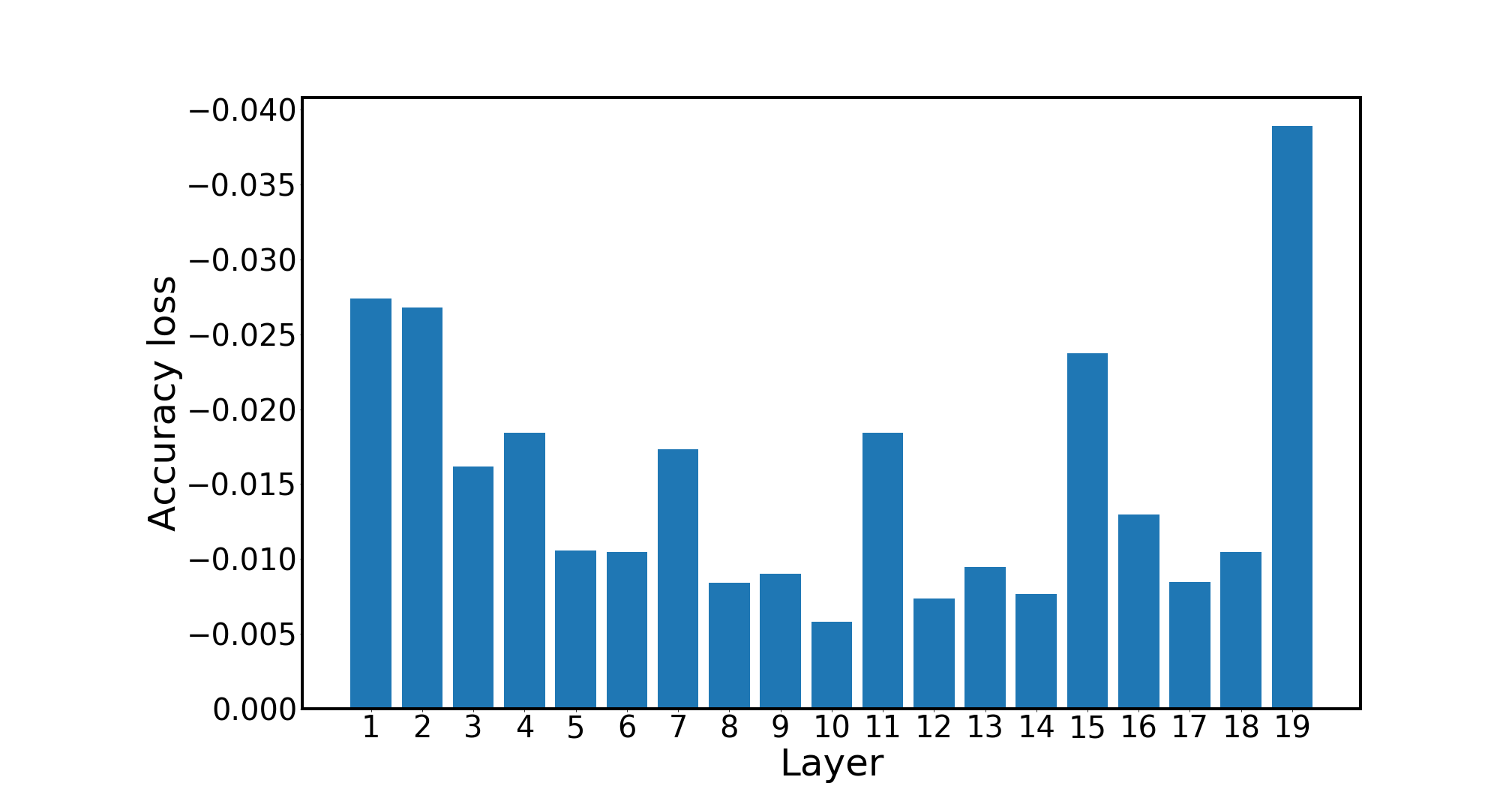}
	\caption{The Maximum accuracy losses in different layers. We argue that they can be the indicators to decide the block to be replaced first.}
	\label{fig:layer_importance}
\end{figure}

Fig.~\ref{fig:layer_importance} shows the largest accuracy losses $\max_i (\boldsymbol{t}_i^{(j)})$ at different layers. As can be seen, different layers learn different information and yield varied accuracy losses. For the layer with smaller accuracy loss, performances of different candidate blocks in this layer are similar. Therefore, when we do network specialization, we prefer to replace $B_G$ by other smaller blocks so that less resource is needed but the validation accuracy just decreases slightly. 

The optimization problem mentioned in Eqn.~(\ref{eq:ga}) is solved by the genetic algorithm. We represent an architecture sampled from the supernet as a chromosome of length 19. A chromosome $(g^{(1)}, g^{(2)}, ..., g^{(19)})$ denotes the block indices of the sub-network constituted by $(B_{g^{(1)}}^{(1)}), B_{g^{(2)}}^{(2)}), ..., B_{g^{(19)}}^{(19)})$. We randomly initialize 20 chromosomes in the first population. The cost of each chromosome is calculated based on the accuracy loss table $T_d$, i.e., $\sum_{l=1}^{19} T_d(l, g^{(l)})$. The top 10 chromosomes that yields the least accuracy loss are selected and grouped into 5 pairs, which are viewed as parent chromosomes. For each pair, a position from 1 to 19 (length of a chromosome) is randomly selected for the crossover operation. Five pairs of children are generated after crossover. For each of this child chromosome, the index of each position may randomly mutate to another index with probability 0.1. After crossover and mutation, these 10 chromosomes and the 10 parent chromosomes form the second-generation population. Notice that, after each crossover and mutation, the resultant chromosome is checked if it fits the constraint $C$. If not, another crossover or mutation operation will be conducted to make new chromosomes. In this work, the same evolution process iterates for 1,000 generations, and the best chromosome in the whole process is picked to represent the architecture of the desired specialized network. 

\section{Experiments}
\subsection{Experimental Settings}
\subsubsection{Datasets.}
We perform all experiments based on the ImageNet dataset \cite{imagenet_cvpr09}. Same as the settings in previous works \cite{chu2019fairnas}\cite{1904.00420}\cite{cai2018proxylessnas}\cite{Tan_2019_CVPR}, we randomly sample 50,000 images (50 images for each class) from the training set as our validation set, and the rest is kept as our training set. The original validation set is taken as our test set to measure the final performance of each model. 

\subsubsection{Training Hyperparameters.} \label{subsection:parameter}
We train the meta-network for 50 epochs using batch size 256 and adopt the stochastic gradient descent optimizer with a momentum of 0.9 and weight decay of $4 \times 10^{-5}$, based on data with standard augmentation (random resizing, cropping, and flipping). We set the initial learning rate to be 0.1, and decay 10x at 20 and 40 epochs. For the fine-tuning stage of PONAS, we follow the same strategy as that for meta training. But we only fine tune each layer for 3 epochs and set the initial learning rate to be 0.001 without any decay setting. 

After determining the architecture of a specific network that conforms to the given constraint, we train the specific network using the standard SGD optimizer with Nesterov momentum 0.9 and weight decay $4e^{-5}$. The initial learning rate is 0.045, and we use the cosine scheduler \cite{1608.03983} for learning rate decay. We use 4 NVIDIA GTX 1080Ti GPUs for training.

\subsection{ImageNet Classification}
Three specific models, named as PONAS-A, PONAS-B, and PONAS-C, are specialized from the supernet according to the accuracy loss table to meet different requirements. It is worth noting again that we only need to search the accuracy loss table and require very little deploy time for finding specific networks. The result is shown in Table~\ref{table:imagenet}. As can be seen, the proposed PONAS requires similar search time to the most recent weight sharing approaches  \cite{chu2019fairnas}\cite{cai2018proxylessnas}\cite{Wu_2019_CVPR}. However, thanks to the design of the accuracy loss table, we achieve the one-shot property \cite{1904.00420}\cite{chu2019fairnas}\cite{cai2020once}, and the time for deploy can be almost ignored. Overall, the top-1 accuracies are 74.67\%, 74.95\%, and 75.2\% for PONAS-A, -B, and -C, respectively, which are quite comparable with the state of the arts (FairNAS \cite{chu2019fairnas}). 

\begin{table}[]
\caption{ImageNet classification performance comparison with the SOTA models. }
\resizebox{\textwidth}{!}{%
\begin{tabular}{l|ccccc|cc|cc}
\hline
Model &
  \begin{tabular}[c]{@{}c@{}}Search\\ method\end{tabular} &
  \begin{tabular}[c]{@{}c@{}}Search\\ space\end{tabular} &
  \begin{tabular}[c]{@{}c@{}}Search\\ dataset\end{tabular} &
  \begin{tabular}[c]{@{}c@{}}Search\\ GPU hours\end{tabular} &
  \begin{tabular}[c]{@{}c@{}}Deploy\\ GPU hours\end{tabular} &
  Params(M) &
  FLOPs(M) &
  \begin{tabular}[c]{@{}c@{}}Top-1\\ acc(\%)\end{tabular} \\ \hline
MobileNetV2 \cite{Sandler_2018_CVPR}        & manual   & -     & -        & - & - & 3.4 & 300 & 72.0 \\
MobileNetV2(1.4X) \cite{Sandler_2018_CVPR}  & manual   & -     & -        & - & - & 6.9 & 585 & 74.7 \\
ShuffleNetV2(1.5X) \cite{Ma_2018_ECCV} & manual   & -     & -        & - & - & 3.5 & 299 & 72.6  \\ \hline
PNASNet \cite{Liu_2018_ECCV}            & SMBO     & Cell  & CIFAR-10 & - & - & 5.1 & 588 & 74.2 \\
DPP-Net-Panacea \cite{Dong_2018_ECCV}    & SMBO     & Cell  & CIFAR-10 & - & - & 4.8 & 523 & 74.02\\
DARTS \cite{liu2018darts}              & gradient & Cell  & CIFAR-10 &  96 & 96$N$  & 4.7 & 574 & 73.3 \\ \hline
FBNet-A \cite{Wu_2019_CVPR}            & gradient & layer & ImageNet &  216 & 216$N$  & 4.3 & 249 & 73.0  \\
FBNet-B \cite{Wu_2019_CVPR}           & gradient & layer & ImageNet & 216  & 216$N$ & 4.5 & 295 & 74.1  \\
FBNet-C \cite{Wu_2019_CVPR}           & gradient & layer & ImageNet & 216  & 216$N$ & 5.5 & 375 & 74.9 \\
SinglePath NAS \cite{stamoulis2019singlepath}  & gradient & layer & ImageNet & 30  & 30$N$  & 4.3  & 365 & 75.0   \\
SinglePath OneShot \cite{1904.00420}      & evolution & layer & ImageNet & 288  & 24$N$  & -   & 328 & 74.7   \\
OFA w/ PS \cite{cai2020once}          &  evolution & layer & ImageNet &   1200 & 40 & -  & 230 & 76.00  \\
ProxylessNAS-R \cite{cai2018proxylessnas}     & RL       & layer & ImageNet & 200 & 200$N$  & 4.1 & 320 & 74.6 \\
MnasNet-A1 \cite{Tan_2019_CVPR}         & RL       & stage & ImageNet &40K & 40K$N$  & 3.9 & 312 & 75.2 \\
MnasNet-A2 \cite{Tan_2019_CVPR}        & RL       & stage & ImageNet & 40K & 40K$N$  & 4.8 & 340 & 75.6 \\
MobileNetV3-Large \cite{Howard_2019_ICCV}  & RL       & stage & ImageNet & 40K & -  & 5.4 & 219 & 75.2  \\
MobileNetV3-Small \cite{Howard_2019_ICCV} & RL       & stage & ImageNet & 40K & - & 2.9 & 66  & 67.4  \\
FairNas-A \cite{chu2019fairnas}           & RL       & layer & ImageNet & 240  & 48& 4.4 & 321 & 74.69  \\ 
FairNas-B \cite{chu2019fairnas}         & RL       & layer & ImageNet & 240  & 48 & 4.5 & 345 & 75.10  \\
FairNas-C \cite{chu2019fairnas}        & RL       & layer & ImageNet & 240  & 48  & 4.6 & 388 & 75.34 \\ \hline

%\textbf{PONAS-A}    & PONAS & layer & ImageNet & 210 & \sim \textbf{0} & 4.6 & 326 & 74.4  \\
\textbf{PONAS-A}    & evolution & layer & ImageNet & 210 & $\sim$ \textbf{0} & 5.1 & 326 & 74.67  \\
% \textbf{PONAS-B}    & PONAS & layer & ImageNet & 210 & \sim \textbf{0} & 4.8 & 342  &  74.7      \\
\textbf{PONAS-B}    & evolution & layer & ImageNet & 210 & $\sim$ \textbf{0} & 5.1 & 349 & 74.95  \\
% \textbf{PONAS-C}    & PONAS & layer & ImageNet & 210 & \sim \textbf{0} &  5.2   & 377 & 75.1  \\  
\textbf{PONAS-C}    & evolution & layer & ImageNet & 210 & $\sim$ \textbf{0} & 5.6 & 376 &  75.2 \\
\end{tabular}
}
\label{table:imagenet}
\end{table}

\begin{figure}
	\centering
	\includegraphics[width=\textwidth]{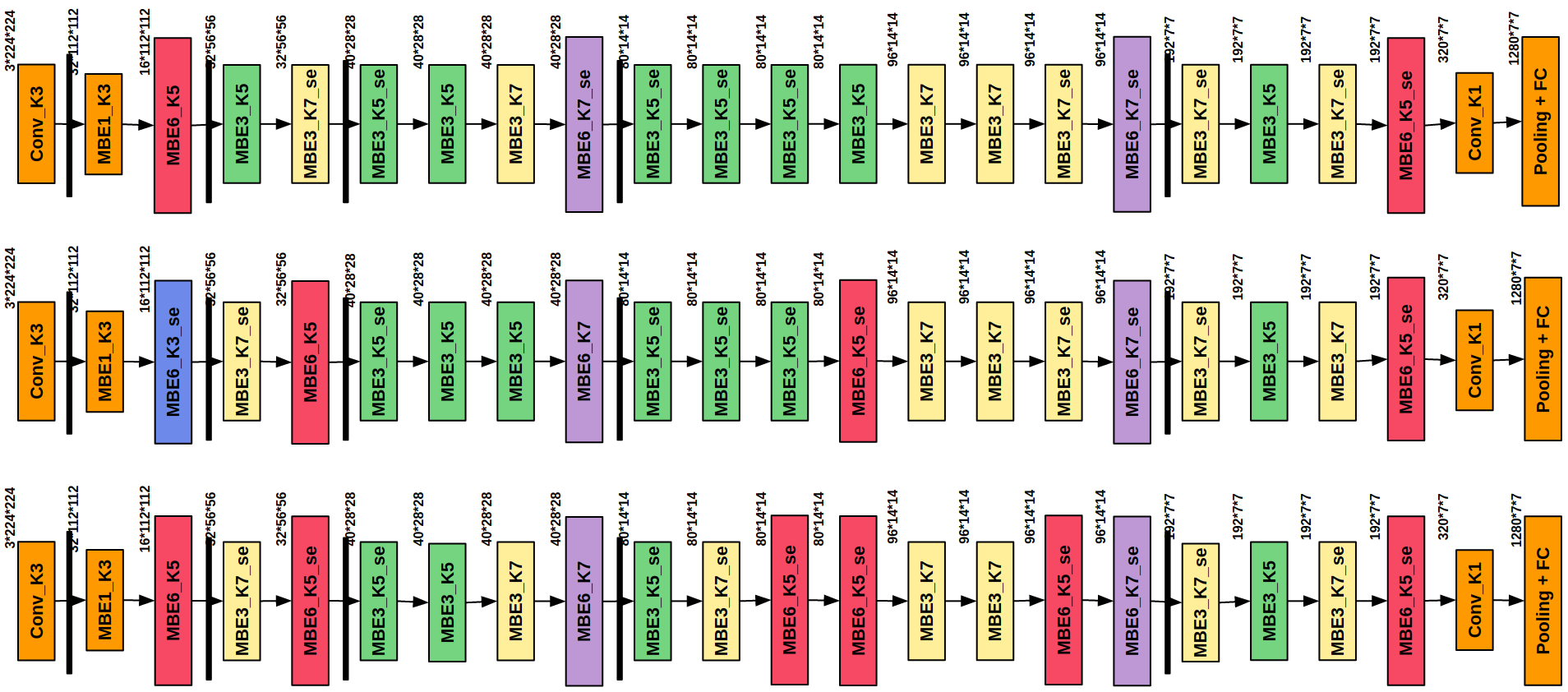}
	\caption{Architectures of PONAS-A, -B, -C from top to down. \enquote{MBE3} and \enquote{MBE6} denote the mobile inverted bottleneck convolution layers with expansion ratios 3 and 6, respectively. \enquote{K$X$} denotes the kernel size of $X$. \enquote{se} denotes whether the  squeeze-and-excite \cite{hu2018senet} module is used. The orange blocks are predefined blocks before searching.}
	\label{fig:architecture}
\end{figure}

Fig.~\ref{fig:architecture} shows architectures of PONAS-A, PONAS-B, and PONAS-C. Notice that the three models tend to choose high expansion rates and large kernel at more important layers (the layers with larger accuracy losses, see Fig.~\ref{fig:layer_importance}), which tends to improve performance. On the other hand, to reduce computational complexity, the three models tend to choose the block with small expansion rate at less important layers. 

\begin{figure}
	\centering
	\includegraphics[width=2.3in]{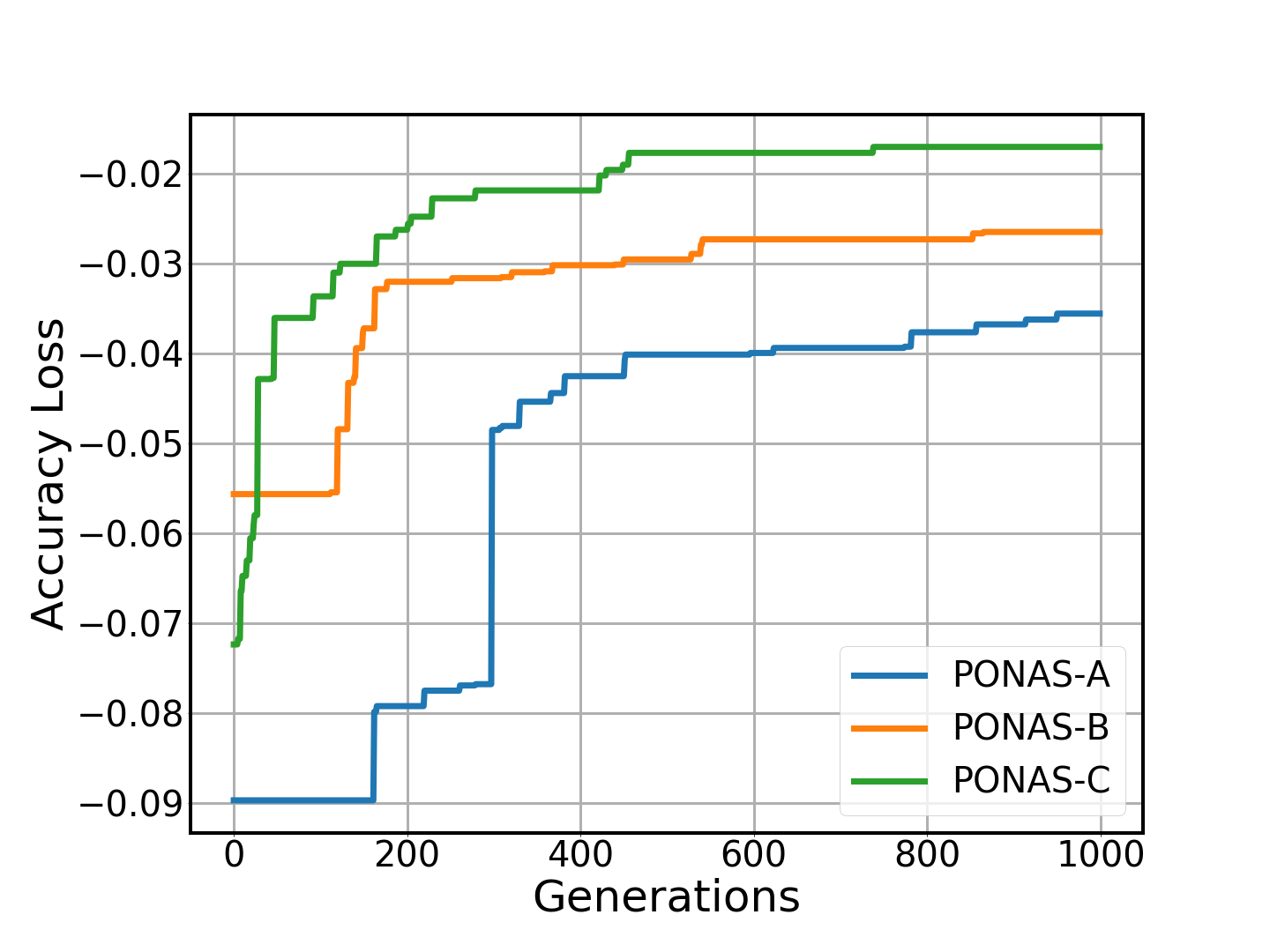}
	\caption{Evolution curves of the network specialization process for finding architectures of PONAS-A, -B, and -C.}
	\label{fig:evolution}
\end{figure}

Fig.~\ref{fig:evolution} shows the evolution curves of finding architectures of three specific models, PONAS-A, PONAS-B, and PONAS-C. Thanks to the accuracy table, we can evolve more generations to get the network with lower accuracy loss. 

\subsection{Ablation Studies}
\subsubsection{Efficiency of Two-stage Training.}
For fair comparison of the proposed two-stage training and FairNAS \cite{chu2019fairnas}, which directly trained the entire supernet from random initialization with strict fairness, we train both methods for 50 epochs and set the initial learning rate as 0.1, and decay 10x at 20 and 40 epochs. For two-stage training, we train 30 epochs for the meta training stage and another 20 epochs for the fine-tuning stage. Instead of progressively fine tuning each layer mentioned in Sec.~\ref{sec:twostage}, we fine tune the entire supernet for fair comparison with FairNAS. The reason to compare with FairNAS is that we both train the supernet with the strict fairness property \cite{chu2019fairnas}. 

% The left of Fig. \ref{fig:two_stage_acc} shows the relationship between validation accuracy and training epochs. %For the proposed two-stage training, the first 30 epochs is the meta training stage, and the latter 30 epochs is the fine-tuning stage. 
% Because the architecture of the meta network is simple, the validation accuracy is boosted rapidly in the meta training stage, reaching up to 70\% after 30 epochs. However, FairNAS converges relatively slower  because of the complex architecture of the supernet. For the proposed two-stage training, the validation accuracy drops at the beginning of the fine-tuning stage but rises quickly after a few epochs. After 60 epochs, the proposed two-stage training reaches up to 70\%, which is 10\% higher than FairNAS. 

\begin{figure}[t]
    \centering
    \subfigure[]{
        \begin{minipage}[t]{0.5\linewidth}
        \centering
         \raisebox{-0.5\height}{\includegraphics[width=2.5in]{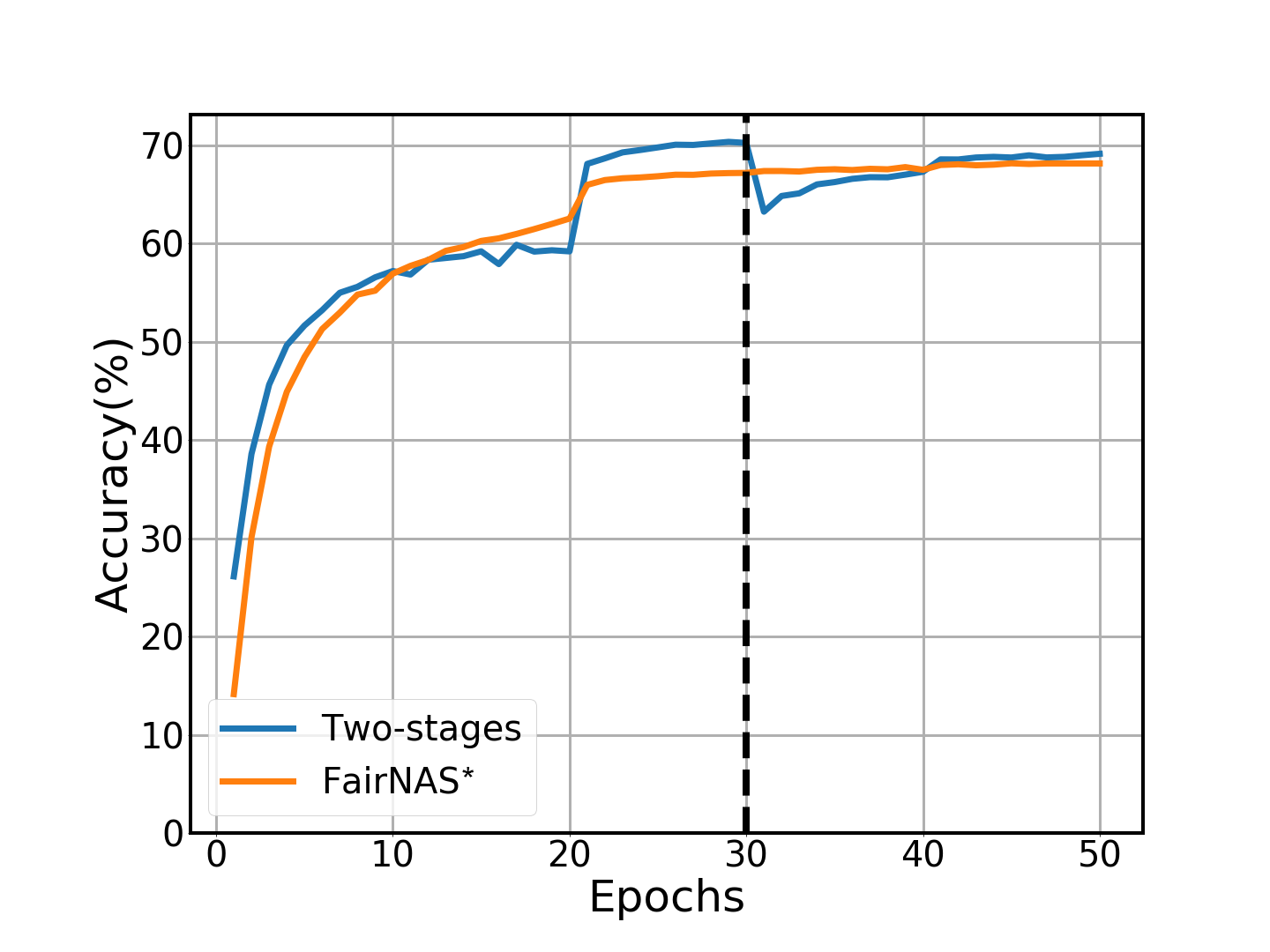}}
        \end{minipage}%
    \label{fig:evo_accuracy}
    }%
    \subfigure[]{
        \begin{minipage}[t]{0.5\linewidth}
        \centering
         \raisebox{-0.5\height}{\includegraphics[width=2.5in]{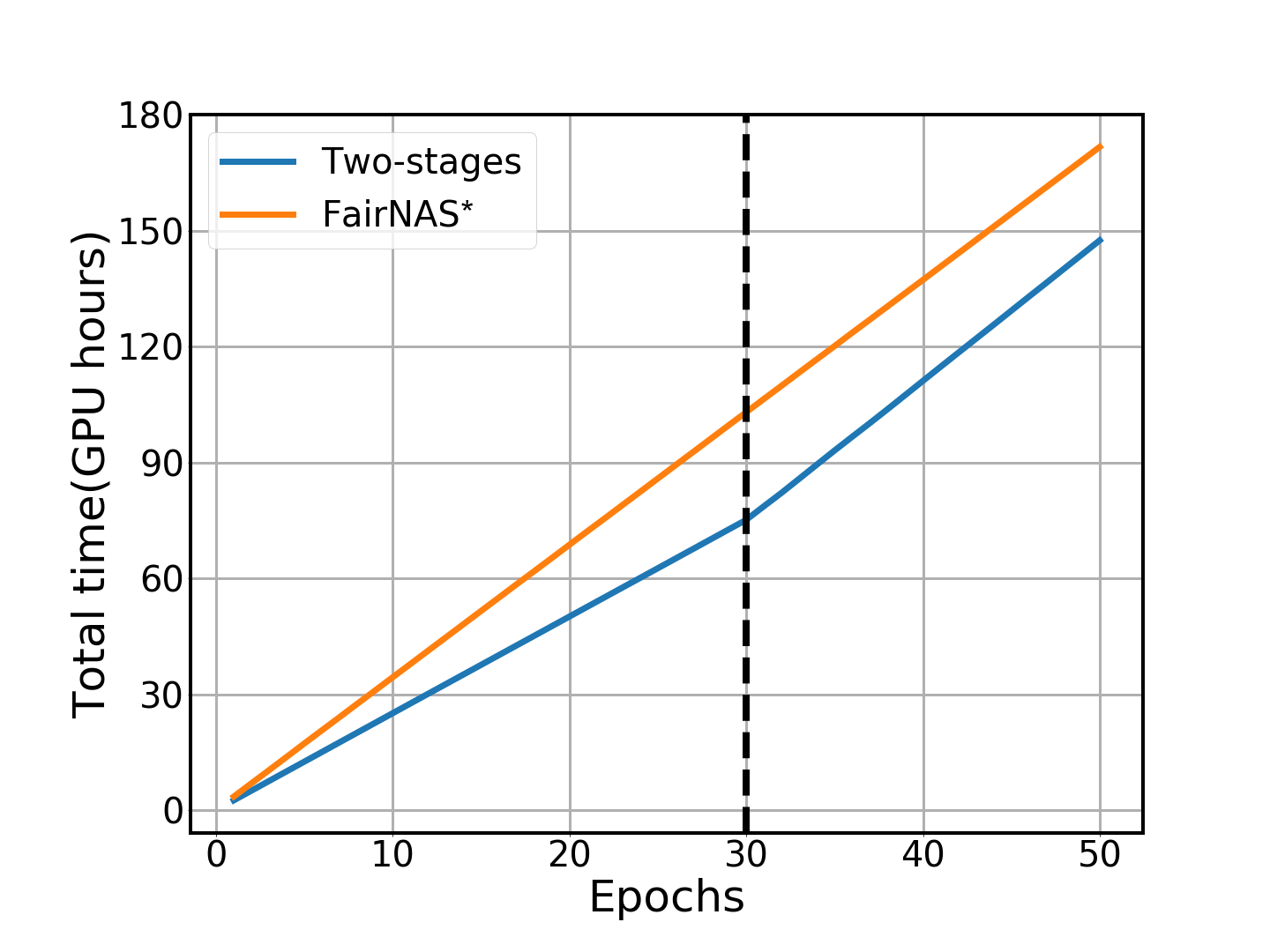}}
        \end{minipage}%
    \label{fig:evo_time}
    }%
    \caption{Comparison between the two-stage training and FairNAS \cite{chu2019fairnas}. The black dash line denotes the beginning of the fine-tuning stage in two-stage training. (a) The evolution of validation accuracies of two methods. (b) The total time required to train the supernets in the manners based on two-stage training and random initialization used in FairNAS. We denote FairNAS with the symbol $^*$ because these results are based on our re-implementation.}
 	\label{fig:two_stage_acc}
    \centering
\end{figure}

Fig.~\ref{fig:evo_accuracy} shows the relationship between top-1 validation accuracy and training epochs. Basically both methods get increasing accuracy as the number of training epoch increases. For the proposed two-stage training, the validation accuracy drops at the beginning of the fine-tuning stage but rises gradually after a few epochs. After 50 epochs, the proposed two-stage training reaches up to 69\%, which is 1\% higher than FairNAS. Fig.~\ref{fig:evo_time} shows comparison in terms of total GPU hours required for training a supernet. In the meta training stage, the two-stage training scheme is 1.35 times faster than because of the simple architecture of the meta network.

% \begin{figure}[t]
% 	\centering
% 	\includegraphics[width=3.7in]{image/two_stage_acc.png}
% 	\caption{Comparison between the two-stage training and FairNAS \cite{chu2019fairnas}. The black dash line denotes the beginning of the fine-tuning stage in two-stage training. Left: The evolution of validation accuracies of two methods. Right: The total time required to train the supernets in the manners based on two-stage training and random initialization used in FairNAS. }
% 	\label{fig:two_stage_acc}
% \end{figure}

\subsubsection{Analysis of the Accuracy Loss Domain.}
\begin{figure}
\begin{floatrow}
\ffigbox{%
	\centering
	\includegraphics[width=2.4in]{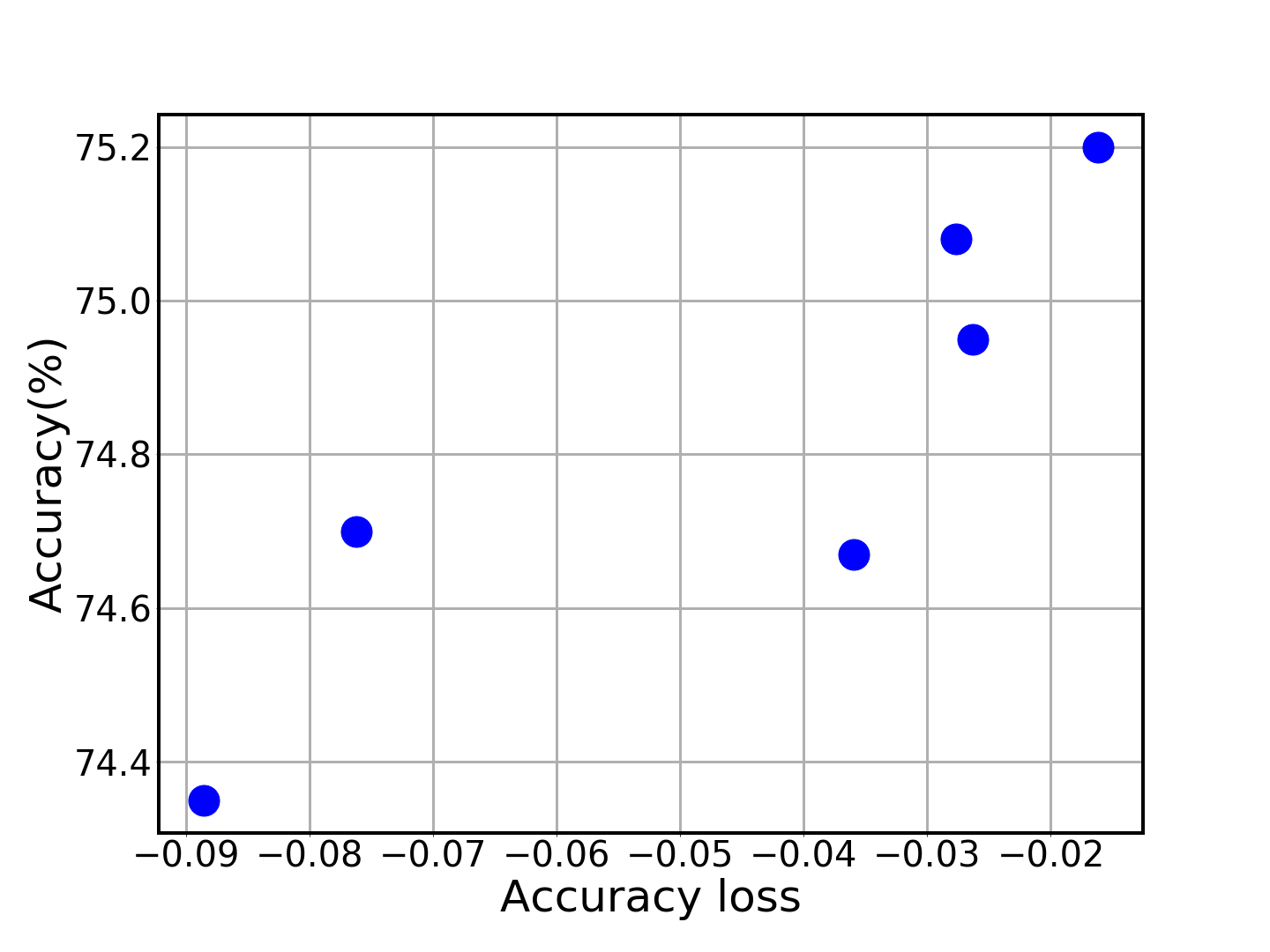}
	\caption{The relationship between validation accuracies of six models and their associated predicted accuracy losses. }%
	\label{fig:loss_accuracy}
}{%
  
}
\ffigbox{%
    \centering
	\includegraphics[width=2.6in]{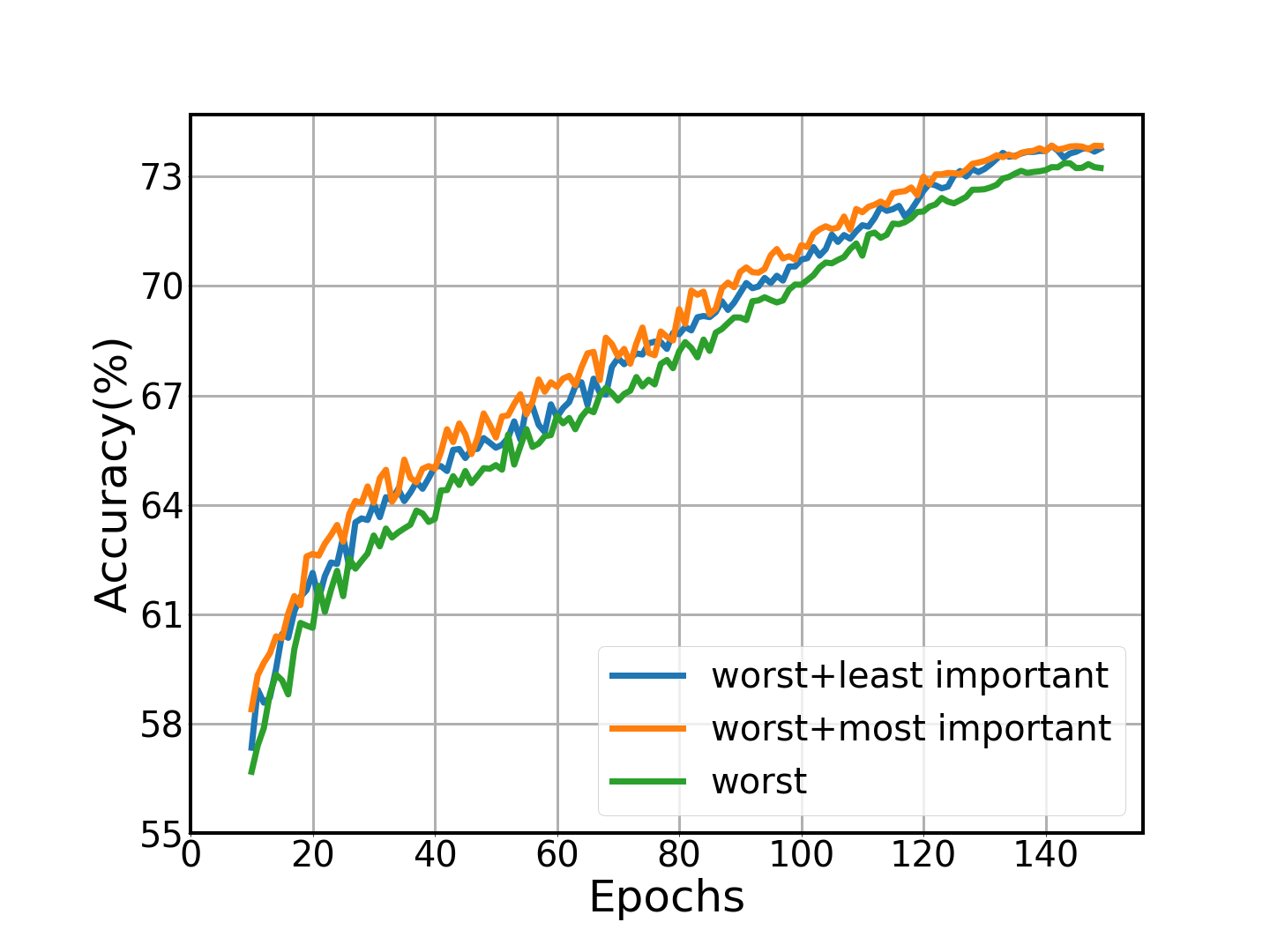}
	\caption{The top-1 validation accuracies of three different networks.}
	\label{fig:two_acc_curve}
}{%
  
}
\end{floatrow}
\end{figure}

We purposely sample six different architectures, train six specialized networks from scratch, and then evaluate them. Fig.~\ref{fig:loss_accuracy} shows the relationship between an architecture's accuracy loss (from the table $T_d$) and the real validation accuracy. We see that these two factors positively correlated. Inspired by \cite{1902.08142}, we adopt the Kendall rank correlation coefficient \cite{10.1093/biomet/30.1-2.81} to measure the correlation. According to the values in Fig.~\ref{fig:loss_accuracy}, the Kendall's $\tau$ value is $\tau=0.733$. This means the architecture with less predicted accuracy loss really yields higher validation accuracy, and verifies the assumption we made in Sec.~\ref{sec:accuracyloss}. 

%We also adopt Kendall Tau \cite{10.1093/biomet/30.1-2.81} for the ranking analysis following a recent work \cite{1902.08142} which evaluates NAS approachs to measure the ranking relation between accuracy loss and the accuracy trained from scratch. The range of $\tau$ is from -1 to 1, meaning the rankings are totally reversed or completely preserved, whereas 0 means there is no correlation at all.We show the ranking comparison with other one-shot method\cite{1904.00420}\cite{pmlr-v80-bender18a}\cite{chu2019fairnas} implemented by \cite{chu2019fairnas} in Table \ref{table:tau}. Accuracy loss achieve the Kendall rank correlation coefficient $\tau=0.733$. It is worth noting that we store the accuracy loss in a table instead of calculating the validation accuracy of the models sampled from supernet\cite{1904.00420}\cite{pmlr-v80-bender18a}\cite{chu2019fairnas}. This make we can deploy various models very efficient.

% \begin{table}
% \begin{floatrow}
% \capbtabbox{%
%     \caption{The ranking ability comparison with other one-shot method\cite{1904.00420}\cite{pmlr-v80-bender18a}\cite{chu2019fairnas} implemented by \cite{chu2019fairnas}.}%
%     \label{table:tau}
%     \begin{tabular}{|c|c|c|}
%     \hline
%     Methods & $\tau$ & \begin{tabular}[c]{@{}c@{}}Deploy  \\ GPU hours\end{tabular} \\ \hline
%     One-Shot\cite{pmlr-v80-bender18a} & 0 & - \\
%     SPOS\cite{1904.00420} & 0.615 & 30$N$ \\
%     FairNAS\cite{chu2019fairnas} & 0.949 & 48 \\
%     \textbf{Accuracy Loss} & 0.733 & $\sim \textbf{0}$ \\ \hline
%     \end{tabular}

% }{%

% }
% \capbtabbox{%
%     \caption{Impact of different layers on performance. The network \emph{most importance} achieves the highest accuracy with the least FLOPs.}
%     \label{table:importance}
%     \begin{tabular}{|c|c|c|}
%     \hline
%     Model & FLOPs & \begin{tabular}[c]{@{}c@{}}Top-1 \\ accuracy (\%) \end{tabular} \\ \hline
%     worst & 279 &  73.4\\

%     \begin{tabular}[c]{@{}c@{}}least importance \end{tabular} & 305 & 73.8 \\

%     \begin{tabular}[c]{@{}c@{}}most importance \end{tabular} & \textbf{295} &  \textbf{73.9}\\ \hline
%     \end{tabular}
% }{%

% }
% \end{floatrow}
% \end{table}
\begin{table}[]
    \centering
    \caption{Impact of different layers on performance. The network \emph{most importance} achieves the highest accuracy with the least FLOPs.}
    \label{table:importance}
    \begin{tabular}{|c|c|c|}
    \hline
    Model & FLOPs(M) & \begin{tabular}[c]{@{}c@{}}Top-1 \\ accuracy (\%) \end{tabular} \\ \hline
    worst & 279 &  73.4\\

    \begin{tabular}[c]{@{}c@{}}worst+least importance \end{tabular} & 305 & 73.8 \\

    \begin{tabular}[c]{@{}c@{}}worst+most importance \end{tabular} & \textbf{295} &  \textbf{73.9}\\ \hline
    \end{tabular}
\end{table}

\subsubsection{Importance of Different Layers.}
To verify the assumption that \emph{the accuracy loss of candidate blocks can quantify importance of candidate blocks in different layers}, we specialize three models from the supernet. First, for each layer of the supernet, we purposely replace the block with one candidate block that yields the largest accuracy loss. This model is called the \emph{worst specific network}. Second, we then purposely replace the block in the least important layer of the worst specific network with the best candidate block in that layer to construct the network called \emph{worst+least importance}. Third, we also replace the block in the most important layer of the worst specific network with the best candidate block in that layer, and obtain a new specific network called \emph{worst+most importance}.  For fair comparison, we actually replace the two least important layers in \emph{worst+least importance} to achieve similar FLOPs with \emph{worst+most importance}. 

We train the three models for 150 epochs under the same setting in Sec.~\ref{subsection:parameter} and show the result in Table~\ref{table:importance}. As expected, the worst specific network obtains the worst performance. When the worst network is transformed to \emph{worst+least importance}, top-1 accuracy boosts to 73.8\%. When the worst network is transformed to \emph{worst+most importance}, the highest accuracy (73.9\%) with less FLOPs can be obtained. The differences on accuracy and FLOPs show the impact of importance of different layers. Fig.~\ref{fig:two_acc_curve} shows the evolution of top-1 validation accuracy of these networks. The \emph{worst+most importance} model has higher convergence speed than other networks. 

In summary, using better blocks at more important layers is more efficient than that at less important layers. From both Table~\ref{table:importance} and Fig.~\ref{fig:layer_importance}, we found that layers with different numbers of channels for input and output have larger accuracy loss and are more impactful to final performance. This observation shows that \emph{the accuracy loss of candidate blocks can quantify importance of candidate blocks in different layers}.

% \begin{table}[]
% \centering
% \caption{Impact of different layers on performance. The network \emph{worst + most importance} achieves the highest accuracy with the least FLOPs.}
% \begin{tabular}{|c|c|c|}
% \hline
% Model & FLOPs & \begin{tabular}[c]{@{}c@{}}Top-1 accuracy (\%) \end{tabular} \\ \hline
% worst & 279 &  73.4\\
% worst + least importance & 305 & 73.8 \\
% worst + most importance & \textbf{295} &  \textbf{73.9}\\ \hline
% \end{tabular}
% \label{table:importance}
% \end{table}

% \begin{figure}
% 	\centering
% 	\includegraphics[width=2.8in]{image/two_acc_curve.png}
% 	\caption{The evolutions of top-1 validation accuracy of three different networks.}
% 	\label{fig:two_acc_curve}
% \end{figure}

\section{Conclusion}
We present a progressive one-shot neural architecture search method which searches the best block in each layer progressively to construct the supernet. When constructing the supernet, we also build a table called \emph{accuracy table} to store the validation accuracy of each candidate block in each layer. By transforming the accuracy table into the accuracy loss domain, candidate blocks in different layers are comparable, and importance of different layers can be measured. Given a constraint, we can simply check the accuracy table to see performance of various specialized architectures and find a specific architecture in around 10 seconds. This is much more efficient than previous one-shot NAS. To speed up and stabilize supernet training, we propose a two-stage training approach, including the meta training stage and the fine-tuning stage. We demonstrate that two-stage training is more stable and converges faster than previous approaches that train the supernet from random initialization. %To achieve network specialization, we transform performance of candidate blocks into the \enquote{accuracy loss} space, i.e., representing performance as the \emph{accuracy loss from the best block at the corresponding layer}. In this way, candidate blocks in different layers are comparable, and importance of different layers can be measured. 
In the evaluation, we show that the proposed PONAS can achieve the state-of-the-art performance with very low-cost deployment. %To the best of our knowledge, we are first to explore that different layers have different importance to model performance. 

\bibliographystyle{splncs04}
\bibliography{egbib}
\end{document}